\documentclass{article}

\usepackage{PRIMEarxiv}

\usepackage[utf8]{inputenc} 
\usepackage[T1]{fontenc}    
\usepackage{hyperref}       
\usepackage{url}            
\usepackage{booktabs}       
\usepackage{amsfonts}       
\usepackage{nicefrac}       
\usepackage{microtype}      
\usepackage{lipsum}
\usepackage{fancyhdr}       
\usepackage{graphicx}       
\graphicspath{{media/}}     
\usepackage{amsmath}
\usepackage{subcaption}
\title{Knee or ROC 
}

\author{
  Veronica Wendt, Byunggu Yu, Caleb Kelly, Junwhan Kim  \\
  Dept. of Computer Science and Information Technology \\
  University of the District of Columbia \\
  Washington, DC USA\\
  \texttt{\{veronica.wendt, byu, caleb.kelly, junwhan.kim\}@udc.edu }\\
}

\begin{document}
\maketitle

\begin{abstract}
Self-attention transformers have demonstrated accuracy for image classification with smaller data sets. However, a limitation is that tests to-date are based upon single class image detection with known representation of image populations. For instances where the input image classes may be greater than one and test sets that lack full information on representation of image populations, accuracy calculations must adapt. The Receiver Operating Characteristic (ROC) accuracy thresh-old can address the instances of multi-class input images. However, this approach is unsuitable in instances where image population representation is unknown. We consider calculating accuracy using the knee method to determine threshold values on an ad-hoc basis. Results of ROC curve and knee thresholds for a multi-class data set, created from CIFAR-10 images, are discussed for multi-class image detection.
\end{abstract}


\section{Introduction}
Self-attention transformers have crossed domains from natural language processing~\cite{vaswani2023attention} to computer vision~\cite{dosovitskiy2021image}. With both increased efficiencies and scalability, trans-former architectures are able to train on models with very large data sets and parameters. Further research has sought to decrease the requirement for large data sets, resulting in the development of compact transformer~\cite{hassani2022escaping} architectures that demonstrate notable image classification accuracy for smaller data sets by adding convolutional tokenization. Research tests to-date are based on single class image detection with known representation of image populations. For instances where the input image classes may be greater than one, determining the accuracy threshold using a Receiver Operating Characteristic (ROC) curve~\cite{gneiting2018receiver} may be sufficient. Yet one of the limitations of the ROC curve is that the accuracy threshold is determined during model training and does not subsequently change once the model is deployed into a live environment. Consequentially, using this approach is unsuitable in a live environment. One possible alternative is to apply the knee method to determine thresholds on an ad-hoc basis.

Image classifiers are the first step in object detection models such as the one-stage model - You Only Look Once (YOLO)~\cite{redmon2016look} and the two-stage model - Faster Regions with Convolutional Neural Networks (Faster R-CNN)~\cite{xie2021oriented}. For example, in YOLO, an image classifier is trained to an acceptable level of accuracy. Once trained, a custom output detection module replaces the original output layer and then again trained on the custom detection model. In this paper, we focus on utilizing the self-attention approach for image classification, exploring its application in both YOLO and Faster R-CNN models. This application shows promise in reducing computational requirements while maintaining comparable or improved accuracy. Our approach involves a thorough analysis of the ROC curve and its associated calculations through three distinct methods, aiming to determine optimal threshold values: the ROC threshold and the knee value.

The subsequent sections of the paper are structured as follows: Section 2 provides an overview of the ROC and Knee backgrounds. In Section 3, we describe the details of the three proposed methods. Section 4 covers the experimental setup and results. Finally, we present our conclusions in Section 5.

\section{Background}

\subsection{ROC and Knee}
Traditional ROC curve calculations are for binary classifications, where four possible test predictions outcomes are possible: true positive (TP), true negative (TN), false positive (FP) and false negative (FN). Calculations for the ROC curve are based on the true positive rate (TPR) plotted against the false positive rate (FPR) for a number of different thresholds, where

\begin{equation}
TPR={\frac {TP}{TP+FN}, FPT = \frac{FP}{FP+TN}} \nonumber
\end{equation}

\begin{figure}
  \centering
  \includegraphics[width=0.4\textwidth]{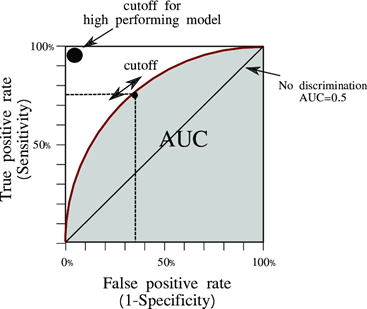} 
  \caption{AUC Example}
  \label{fig:fig1}
\end{figure}

In statistical analysis, the TPR and FPR are also referred to as sensitivity and specificity, respectively. From these calculations and the subsequent plotted ROC curve, the threshold (T) with the greatest area under the curve (AUC) value is obtained at the point where the perpendicular distance between the ROC curve and values of AUC = 0.5 is the greatest. Figure~\ref{fig:fig1} depicts a traditional ROC curve, the value where TPR = 75\% and FPR = 35\% represents the optimal threshold value.

An alternate approach to determining the point of maximum benefit for a system is to calculate the knee value, which is the point of maximum curvature in a data set~\cite{5961514}. The Knee method seeks to find the point in which the relative computing cost of getting more than X correctly labelled images is no longer worth the corresponding benefit. This knee method may be well suited for adaptation in multi-class image classification, as it can be calculated on an ad-hoc basis both during model training and in a live environment, whereas the ROC curve is calculated during model training only.

\subsection{Compact Convolutional Transformer}
Current data sets, such as Fashion MNIST and CIFAR-10~\cite{cifar10}, are useful to a point. They provide small to medium data sets for experimentation with new or adapted image processing methods. One method, compact convolutional transformer (CCT), is a type of self-attention transformer that adds a convolution element, thus combining the advantages of convolution, such as sparse interaction and weight sharing, with the advantages of transformers, such as computational efficiency and scalability~\cite{hassani2022escaping, Khan_2022}. In particular, CCT has demonstrated high levels of accuracy with small data sets at a lower computational cost than vision transformer (VIT) or traditional convolutional neural networks (CNNs)~\cite{hassani2022escaping} with impressive accuracy levels for small scale, small resolution data sets without pre-training. Even so, that research is based upon single image detection with known representation of image population.

\section{Proposed ROC and Knee}
\subsection{Method 1}
For Method 1, the intent is to obtain a maximum threshold value, as represented by $T_{max}$. This is done in several steps.

\begin{figure}
  \centering
  \includegraphics[width=0.5\textwidth]{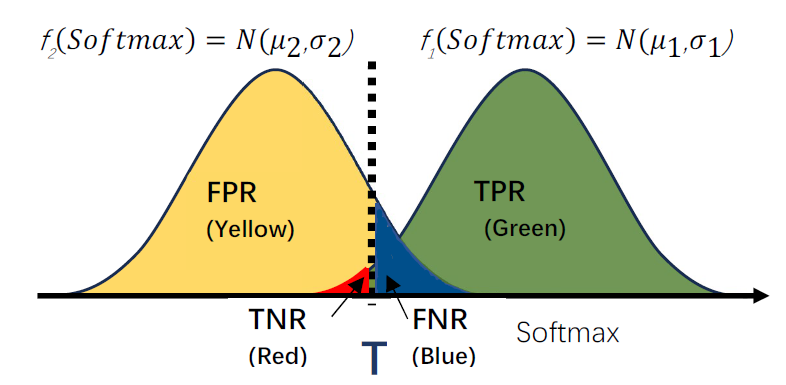}
  \caption{Normal Distribution Curves for TPR and FPT Arrays}
  \label{fig:fig2}
\end{figure}

Step 1: Create two $Softmax$ normal distribution curves based on the values of $TPR(\mu1, \sigma1)$ and $FPR(\mu2,\sigma2)$. Plotting the results depicts two overlapping normal distribution curves as shown in Figure~\ref{fig:fig2}.

Step 2: Chose an arbitrary threshold value, T, anywhere in the domain. In Figure~\ref{fig:fig2}, $T$ is an arbitrary value that divides the normal distribution curves into two distinct areas under the curve.  The $TPR(\mu1, \sigma1)$ areas are represented in green and red, where the green is the true positive rate (TPR) and the red is the true negative rate (TNR). The $FPR(\mu2, \sigma2)$ areas are represented in yellow and blue where the yellow is the false positive rate (FPR) and the blue is the false negative rate (FNR).

Step 3: For the normal distribution curves, we use the cumulative distribution function to approximate the integral values from $T$ to $\infty$. The $T_{max}$ value for the $TPR$ and $FPR$ rates can then be determined using the following equations:
\begin{equation}
\begin{split}
TPR(T_{max}) = CDF(\infty, \mu_{TPR}, \sigma_{TPR}) - CDF( T, \mu_{TPR}, \sigma_{TPR}) \nonumber \\
and  \\
FPR(T_{max}) = CDF(\infty, \mu_{FPR}, \sigma_{FPR}) - CDF( T, \mu_{FPR}, \sigma_{FPR})
\end{split}
\end{equation}

Where $\infty$ can be approximated by using a very large number and CDF is defined as:
\begin{equation}
CDF(T, \mu, \sigma) = \frac{1}{2} [ 1 + erf(\frac{T-\mu}{\sigma \sqrt{2}})] \nonumber
\end{equation}

where $erf$ is the Gauss error function.

The result is a pair of values, one representing the TPR value and one representing the TPN value. This pair of values can now be plotted on a traditional ROC curve.

Step 4: To plot the entire ROC curve, the T value is incremented by a specific value and Step 3 is repeated. The T value continues to increment along a defined range, with Step 3 repeating each increment. The result of all the plotted TPR, FPR values is a novel ROC curve.

\subsection{Method 2}
Method 2 is the second novel approach for obtaining the ROC threshold for a multi-class data set. The intent of Method 2 is to apply a "what if" scenario to rearrange the original arrays (Array1 and Array2) at specified thresholds and then plot a ROC curve based upon these rearranged "new" arrays. In this case, Array1 represents the ground truth positive values, named Positive Array ($P_{arr}$), while Array2 represents the ground truth negative values, named Negative Array ($N_{arr}$). 

\begin{figure}
  \centering
  \includegraphics[width=0.5\textwidth]{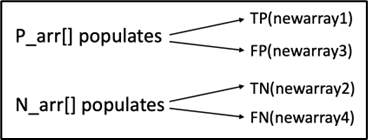}
  \caption{Populating New Array in Method 2}
  \label{fig:fig3}
\end{figure}

To rearrange the arrays, we define four new arrays: True Positives (TP), False Negatives (FN), False Positives (FP) and True Negatives (TN). By following the steps as shown in Figure~\ref{fig:fig3}, the values of $P_{arr}$ and $N_{arr}$ are moved into new arrays. From there, the ROC creation methodology detailed in Method 1 is repeated for Method 2.

The "what if" scenario is done in several steps:
Step 1: Create the following four arrays in preparation for rearranging the Positive and Negative Arrays:

newarray1  = True Positives (TP)  \\
newarray2  = False Negatives (FN)  \\
newarray3  = False Positives (FP)  \\
newarray4  = True Negatives (TN) \\

Step 2: Assign an initial threshold value T and an increment value. In this case, the initial threshold value $T$ = .05 and the increment value is .01.

Step 3: Given an initial threshold value $T$, take $P_{arr}$ and $N_{arr}$, examine each probability value, and assign that value according to the following definitions:

	a) if $P_{arr}$[i] $\geq$ T, 
	    	then newarray1[i] = $P_{arr}$[i] \\
	b) if $N_{arr}$[i] $\geq$ T,  
	    	then newarray2[i] = $N_{arr}$[i] \\
	c) if $P_{arr}$[i] $<$ T,  
	    	then newarray3[i] = $P_{arr}$[i] \\
	d) if $N_{arr}$[i] $<$ T,  
	    	then newarray4[i] = $N_{arr}$[i] \\
where, $i$ represents each iteration.

Step 4: Repeat the same process detailed in Method 1, resulting in a ROC Curve.

Step 5: Increment the threshold value by .01 and repeat Steps 3-4. Once the threshold value equals the maximum threshold value set by the tester, the Method 2 process ends.

Once all ROC curves are plotted, we select the ROC curve from either Method 1 or 2 with the greatest distance from AUC = .5 and obtain the knee value for that ROC curve using the $KneeLocator$ function. Using the knee value, we calculate the maximum threshold value for the best in-class ROC curve.

\subsection{Method 3}

\begin{figure}
  \centering
  \includegraphics[width=0.5\textwidth]{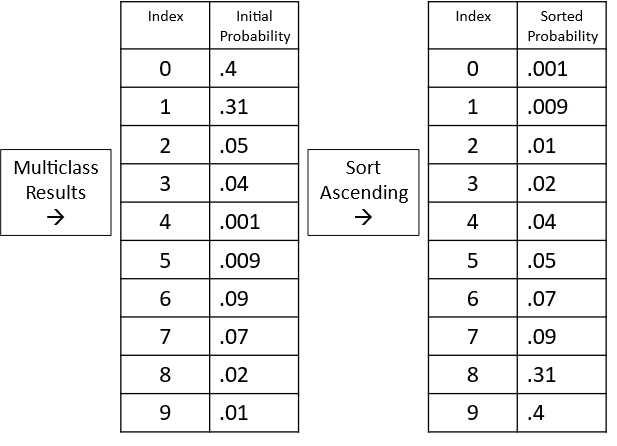}
  \caption{Knee Values Based on Probabilities}
  \label{fig:fig4}
\end{figure}

Method 3, a third approach for evaluating multi-class images, obtains the knee values of a given probability set. This method can be applied on an ad-hoc basis, and thus is suitable for training, testing, and live environments. In this method, with each of the probability values received during the multi-class image testing CCT method, the resultant array is sorted in ascending value and then plotted on a graph, with the probabilities as the y-axis values and index values 0-9 on the x-axis. Figure~\ref{fig:fig4} shows an example of this approach.

The ascending array is then plotted, and a knee value is obtained for each multi-class image.

\section{Experimental Results}

\subsection{Results for Methods 1 and 2}
\begin{figure}
  \centering
  \includegraphics[width=0.4\textwidth]{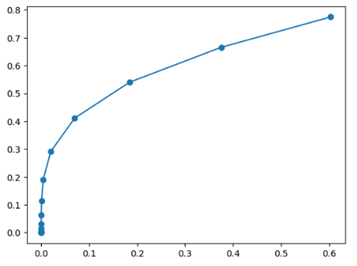}
  \caption{ROC Curve - Initial Threshold using Cumulative Distribution Function (CDF)}
  \label{fig:fig5}
\end{figure}

\begin{figure}
  \centering
  \includegraphics[width=0.4\textwidth]{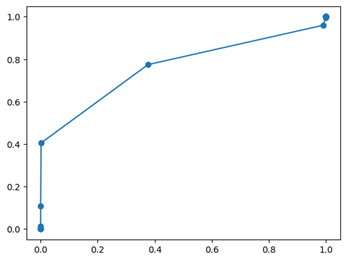}
  \caption{ROC Curve - Sample Variants with minimal population redistribution (Threshold = 0.58)}
  \label{fig:fig6}
\end{figure}

The results of Method 1, where Array1 and Array2 are populated with the multi-class test probabilities, representing the TPR and FPR for the given data set. Figure~\ref{fig:fig5} depicts the ROC curve based upon the initial array values, as calculated using the cumulative distribution function (CDF). The results of Method 2 are found by iterating through different threshold values at a predetermined incremental step of .01. With every iteration, a new ROC curve is generated. Initial iterations do not result in ROC curve generation because a minimum of two data points from the original array must transfer to the new array before enough data is available to generate the ROC curve iteration.  Figure~\ref{fig:fig5} depicts an early iteration of the ROC curve where the threshold value is .58 and the number of new array values for $P_{arr}$ and $N_{arr}$ are both less than 20. Figure~\ref{fig:fig6} demonstrates that a viable ROC curve is generated with the new array values.

\begin{figure}
  \centering
  \includegraphics[width=0.4\textwidth]{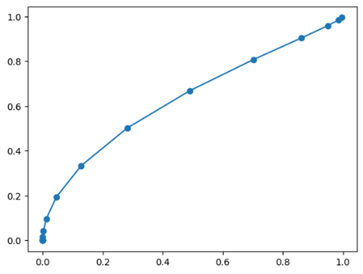}
  \caption{ROC Curve - Sample Variants with 25\% population redistribution (Threshold = 0.20)}
  \label{fig:fig7}
\end{figure}

However, as threshold iterations decrease and array populations redistribute, the resultant ROC curves trend towards near-random values (AUC = .5). Figure~\ref{fig:fig7} depicts a mature iteration of the ROC curve where the threshold value is .20 and the arrays have redistributed 25\% of their population to new array values for $P_{arr}$ and $N_{arr}$. Further iterations of ROC curves result in threshold values that are less than the original ROC curve.

\begin{figure}
    \centering
    \begin{subfigure}{0.4\textwidth}
        \includegraphics[width=\linewidth]{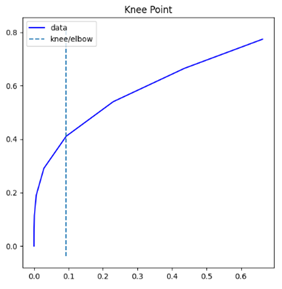}
        \caption{Knee Point of ROC Curve with Largest AUC}
        \label{fig:sub1}
    \end{subfigure}
    \begin{subfigure}{0.4\textwidth}
        \includegraphics[width=\linewidth]{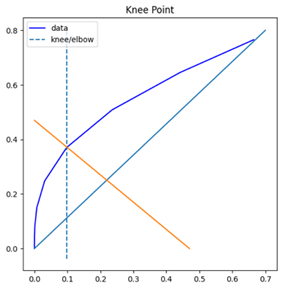} 
        \caption{Calculating Threshold (T) = .693}
        \label{fig:sub2}
    \end{subfigure}
    \caption{Knee Point of ROC and Threshold}
    \label{fig:fig8}
\end{figure}

With all increments calculated, the largest threshold $T$ is selected from the original ROC Curve. We first calculate the knee point with the $KneeLocator$ function to obtain the coordinates of (.098, .371), as shown in Figure~\ref{fig:sub1}. We then calculate the Euclidean distance from the perpendicular point between the knee and where AUC = .5, as shown in Figure~\ref{fig:sub2}. For the knee value of (.098, .371), the AUC coordinates are (.235, .235) and the Euclidean distance is .193. The threshold is then calculated by adding .5 to the distance, resulting in $T$ = .693. This is the maximum threshold for the multi-class image data set.

One possible reason for why none of the Method 2 ROC curves outperform Method 1 is that the valuations for TPR and FPR may require reformulation, where True Negatives and False Negatives are included in the calculations. Another possible reason is that the probability values for the FPR may be so much smaller than the TPRs that the results mimic an imbalanced data set.

\subsection{Results for Method 3}
\begin{figure}
    \centering
    \begin{subfigure}{0.4\textwidth}
        \includegraphics[width=\linewidth]{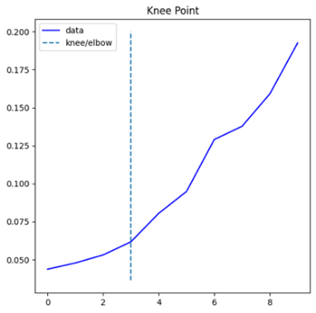}
        \caption{Knee = 3}
        \label{fig9:sub1}
    \end{subfigure}
    \begin{subfigure}{0.4\textwidth}
        \includegraphics[width=\linewidth]{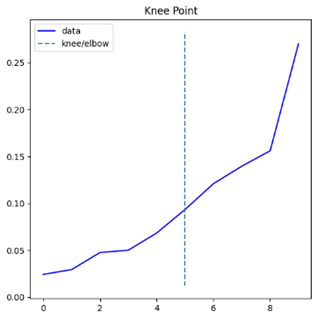} 
        \caption{Knee = 5}
        \label{fig9:sub2}
    \end{subfigure}
    \caption{(a) Knee Point, Max probability = .20 (b) Knee Point, Max probability = .25}
    \label{fig:fig9}
\end{figure}

\begin{figure}
    \centering
    \begin{subfigure}{0.4\textwidth}
        \includegraphics[width=\linewidth]{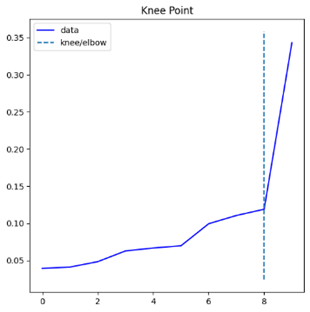}
        \caption{Knee = 8}
        \label{fig10:sub1}
    \end{subfigure}
    \begin{subfigure}{0.4\textwidth}
        \includegraphics[width=\linewidth]{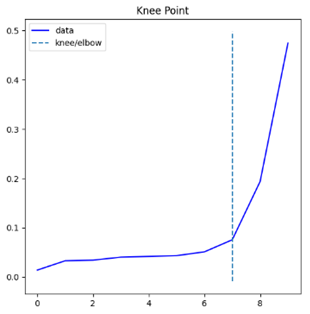} 
        \caption{Knee = 7}
        \label{fig10:sub2}
    \end{subfigure}
    \caption{(a) Knee Point, Max probability = .35 (b) Knee Point, Max probability = .50}
    \label{fig:fig10}
\end{figure}

Method 3, calculating the knee values, yields a result that cannot easily be compared to the results of Methods 1 and 2. Primarily, rather than creating a ROC curve, the results solely depict the knee values for each multi-class image. Figure~\ref{fig:fig9} depicts sample knee values for maximum probability values of (a) .20, (b) .25. Figure~\ref{fig:fig10} depicts sample knee values for maximum probability values of (a) .35, and (b) .50 respectively.

\begin{figure}
  \centering
  \includegraphics[width=0.21\textwidth]{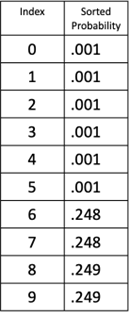}
  \caption{Best Case Probabilities for 4-class image, 10 Classes}
  \label{fig:fig11}
\end{figure}

Results of these knee calculations show a pattern when at least one of the maximum probability values $\geq$  .35, then the knee values are consistently greater than six.  In the sample multi-class data set, 58.8\% of the images (294 of 500) contained a probability value $\geq$ .35. Among the 294 multi-class images, the likelihood that 6 $\leq$ Knee $\leq$ 8 is 92.5\% (272 of 294). For maximum probability values $<$ .35, the knee values were inconsistent and occasionally non-existent. Because there are 10 different probability values for a 4-class image, the best possible case for probability results would be something akin to Figure~\ref{fig:fig11}.

\subsection{Findings}
In method 1, Results in a viable ROC curve that is derived from the cumulative distribution function, as applied to normal distribution curves for TPR and FPR. In method 2, Further iterations of ROC curves in "what if" scenarios, that apply Method 2, result in threshold values that are less than the ROC curve produced by Method 1. In method 3, when images are large, i.e., probability $\geq$ .35, knee values are consistently greater than six. When images are small, i.e., probability $<$ .35, knee values are inconsistent and occasionally non-existent. Based on the above results, Method 1 is appropriate for small multi-class objects and Method 3 is appropriate for large multi-class objects.

\section{Conclusion and Future Remarks}
Self-attention transformer research has focused on image classification for single images with known representations for image populations. For instances where the input image may be greater than one and test sets lack information on the representation of image populations, accuracy calculations must adapt. Our proposed adaptations, to directly calculate the ROC curve accuracy threshold using the cumulative distribution function and to calculate the knee value base on prob-ability outcomes provides a first step at adaptation, with the opportunity to identify more than one class in a multi-class image.

Tackling the research topics of multi-class image detection using new architectures, such as CCT requires multiple iterations of research and experimentation. This paper represents a first step in combining the lightweight computation benefits of transformer-based architectures with the necessity to identify multiple classes in any given image. Future works will look to refine the ROC calculations to validate the TPR and FPR values and repeat the experiment on a larger ImageNet 1000 dataset.  Additionally further research and experiments with knee calculations for multi-class images may result in an incremental step of identifying more than one class. This approach may then be combined with relevant architectures to yield state-of-the-art results. 

Not addressed in this paper is the issue of identifying classes where image population representation is unknown. Further research will identify opportunities to include this element in the architecture calculations, specifically implementing more mature iterations of the ad-hoc knee method.

\section*{Acknowledgments}
This work has been supported by US National Science Foundation CISE-MSI (RCBP-ED: CNS: Data Science and
Engineering for Agriculture Automation) under grant 2131269.

\bibliographystyle{unsrt}  
\bibliography{references}

\end{document}